\documentclass[letterpaper, 10 pt, journal, twoside]{ieeeconf}  % Comment this line out if you need a4paper

\IEEEoverridecommandlockouts                              % This command is only needed if 
\usepackage{amsthm}
\usepackage{multicol}
\usepackage[bookmarks=true]{hyperref}
\usepackage{amsmath}
\usepackage[dvipsnames]{xcolor}
\usepackage{graphicx}
\usepackage[footnotesize]{caption}
\usepackage{subcaption}
\usepackage{amssymb}
\usepackage{array}
\usepackage{url}
\usepackage{color}
\usepackage{nccmath}
\usepackage[normalem]{ulem}
\usepackage[linesnumbered,ruled,vlined]{algorithm2e}\usepackage{enumerate}
\usepackage{multirow}
\usepackage{bbm}
\usepackage{marginnote}
\usepackage[T1]{fontenc}

\usepackage[letterpaper, left=0.667in, right=0.667in, bottom=0.597in, top=0.792in]{geometry}

\renewcommand{\theenumi}{\roman{enumi}}%
\SetCommentSty{mycommfont}

\SetKwInput{KwInput}{Initialize}
\SetKwInput{KwOutput}{Output}
\SetKwInOut{Requires}{Requires}
\SetKwInOut{Return}{Return}
\SetKwInOut{Algorithm}{Algorithm}

\newtheorem{defn}{Definition}

\newtheorem{assum}[defn]{Assumption}

\newtheorem{thm}[defn]{Theorem}

\title{\LARGE \bf
These Maps Are Made For Walking: Real-Time Terrain Property Estimation for Mobile Robots
}

\author{Parker Ewen$^{1}$, Adam Li$^{1}$, Yuxin Chen$^{1}$, Steven Hong$^{1}$, and Ram Vasudevan$^{1}$% <-this % stops a space
\thanks{$^{1}$All authors affiliated with the Robotics Institute at the University of Michigan, 2505 Hayward Street, Ann Arbor, Michigan, USA, \{\tt \small pewen, adamli, chyuxin, hongsn, ramv\}{\tt \small @umich.edu}}%
\thanks{This work is supported by the Ford Motor Company via the Ford-UM Alliance under award N022977, by the Office of Naval Research under Award Number N00014-18-1-2575, and in part by the National Science Foundation under Grant 1751093.}%
}

\begin{document}

\maketitle
\thispagestyle{empty}
\pagestyle{empty}

%%%%%%%%%%%%%%%%%%%%%%%%%%%%%%%%%%%%%%%%%%%%%%%%%%%%%%%%%%%%%%%%%%%%%%%%%%%%%%%%
\begin{abstract}
The equations of motion governing mobile robots are dependent on terrain properties such as the coefficient of friction, and contact model parameters.
Estimating these properties is thus essential for robotic navigation.
Ideally any map estimating terrain properties should run in real time, mitigate sensor noise, and provide probability distributions of the aforementioned properties, thus enabling risk-mitigating navigation and planning.
This paper addresses these needs and proposes a Bayesian inference framework for semantic mapping which recursively estimates both the terrain surface profile and a probability distribution for terrain properties using data from a single RGB-D camera.
The proposed framework is evaluated in simulation against other semantic mapping methods and is shown to outperform these state-of-the-art methods in terms of correctly estimating simulated ground-truth terrain properties when evaluated using a precision-recall curve and the Kullback-Leibler divergence test.
Additionally, the proposed method is deployed on a physical legged robotic platform in both indoor and outdoor environments, and we show our method correctly predicts terrain properties in both cases. 
The proposed framework runs in real-time and includes a ROS interface for easy integration.
\end{abstract}


%%%%%%%%%%%%%%%%%%%%%%%%%%%%%%%%%%%%%%%%%%%%%%%%%%%%%%%%%%%%%%%%%%%%%%%%%%%%%%%%
\section{Introduction} \label{sec:intro}
Mapping from images or point clouds enables mobile robots to perform object avoidance and terrain traversal \cite{mastalli2017}.
Prior work in mapping for mobile robots has focused on generating maps by reconstructing the surface geometry in the vicinity of the robot as a 2.5-D polygonal mesh \cite{fankhauser2018}, often referred to as an elevation map or contact surface.
These representations describe the geometry of the robot's surroundings, but have no information regarding the properties of the underlying terrain such as the friction coefficient.
The equations of motion governing the behavior of mobile robots are a function of both the internal state of the robot and the properties of the terrain over which the robot is traversing \cite{neunert2018}. 
As a result, maps used for robot navigation should include information about these properties.
Work has been done in estimating terrain properties such as friction or the internal shear coefficients of granular surfaces from single RGB images \cite{nguyen2021, noh2021, brandao2016}.
Building semantic maps with terrain property estimates using these methods has remained a challenge due to the difficulty of incorporating prior information into these maps.

\begin{figure} [!t]
    \centering
    \includegraphics[width=\columnwidth]{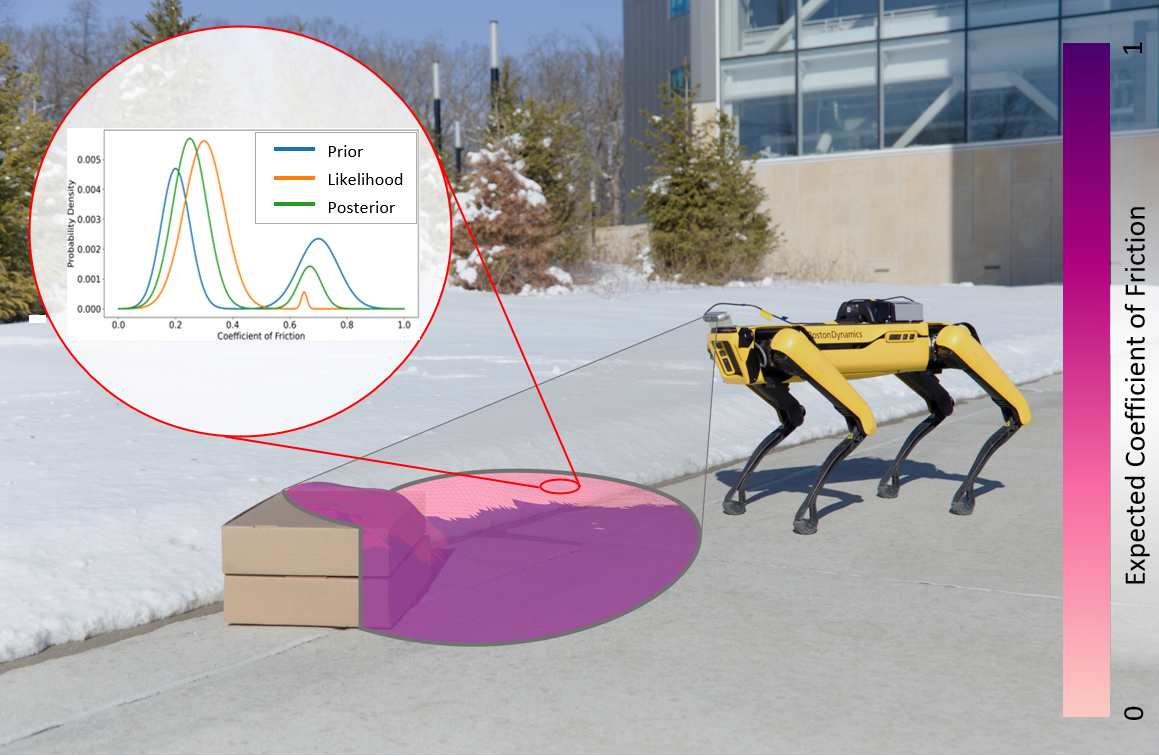}
    \caption{
    An illustration of the real-time semantic mapping method proposed in this paper which recursively estimates terrain height and friction properties from RGB-D images.
    A triangular mesh represents the probabilistic estimate of the contact surface of the robot's surroundings.
    The proposed algorithm estimates terrain classes for each face of the triangular mesh using Bayesian inference and off-the-shelf semantic segmentation networks.
    Probability distributions for terrain properties are then computed (shown in the red inlay for one region).
    This algorithm runs on several robot systems including Boston Dynamics' Spot and Agility Robotics' Digit.
    }
    \label{fig:main_figure}
\end{figure}

As illustrated in Fig. \ref{fig:main_figure}, the contributions of this paper are two-fold.
First, we develop a novel dataset that is used to model the relationship between the coefficient of friction and a variety of terrain classes.
Second, we propose a robot-centric semantic mapping framework by which geometric and terrain properties are estimated using a closed-form Bayesian inference algorithm.

The maps generated by this algorithm map 5m around the robot with discretizations of 2cm and are computed at speeds between $9$Hz $\pm 5$Hz.
The proposed mapping algorithm is accessible via an open-source ROS package which takes RGB-D images and camera pose mean and covariance estimates as inputs and outputs the semantic map.
We use our semantic mapping framework to estimate the coefficient of friction both in simulation and in the real-world and show that it outperforms existing methods from the literature in terms of accuracy in property estimation.

The remainder of this paper is organized as follows:
Section \ref{sec:related} summarizes the mapping literature and Section \ref{sec:prelim} introduces preliminary concepts used throughout the paper.
Section \ref{sec:method} provides an overview of our method and Sections \ref{sec:recursive_elevation_mapping} and \ref{sec:terrain_propety_est} pertain to the recursive elevation mapping and recursive terrain property estimation portions of our algorithm, respectively.
Section \ref{sec:implementation} discusses implementation details and describes our dataset used to model the coefficient of friction for ten terrain classes.
Section \ref{sec:results} describes the evaluation of our algorithm both in simulation and in real-world scenarios.
Section \ref{sec:conclusion} follows with concluding remarks and a discussion on future work.

\section{Related Works} \label{sec:related}
This section describes the existing literature on geometric and semantic mapping methods with an emphasis on algorithms pertaining to legged locomotion.
Geometric maps represent the interface between free-space and occupied space, and are used to model the environment around the robot.
Semantic maps include information such as terrain class, terrain properties, or additional high-level labels within the representation.
Often semantic maps include a geometric representation onto which the semantic information is projected, and thus in these cases, they may contain more information than geometric maps.

\subsection{Geometric Mapping}

A robot's surroundings can be geometrically modelled by constructing representations of the underlying terrain surface using range sensor data.
These range sensors produce sparse point cloud representations, which must then be converted into continuous or piecewise structures to be of use for planning.
We organize prior work in geometric mapping into either volumetric or 2.5-D piecewise-planar categories.

Volumetric representations map the 3D geometry of a scene using discretized volumes, such as voxelized occupancy grids, or via Truncated Signed Distance Fields (TSDF) or Euclidean Signed Distance Fields (ESDF). 
Voxelized occupancy maps, for instance, are the three dimensional equivalents to occupancy grid maps.
Voxels are given a binary label representing either free-space or occupied space, and the probability of each label given new data is updated by applying a Bayes filter.
Such volumetric representations are often memory and computational resource intensive, but one may address some of these limitations by applying memory-efficient representations such as octrees \cite{hornung2013}.
Other representations, like TSDFs and ESDFs,  map the distance to the nearest occupied cell rather than storing the entire volumetric representation of the environment \cite{oleynikova2016}.

Often for legged robotic locomotion and footstep planning, only a lower-dimensional rather than volumetric representation is required.
Point-foot robots, such as Boston Dynamics's Spot, require maps with discretizations on the order of 1cm to perform footstep planning due to their footprint size \cite{fankhauser2018}.
State-of-the-art volumetric representations with 1cm resolution are only able to operate at around 1Hz \cite{hornung2013}.

The most common geometric mapping paradigm in legged robotics is the elevation map, where a piecewise-planar mesh is used to represent the terrain \cite{mastalli2017, fankhauser2018}.
Such maps also have parallels in the surface reconstruction community \cite{zienkiewicz2016}.
Recently, Bayesian inference has been applied to recursively update an elevation map to minimize the impact of noise from the range sensor measurements and robot pose estimation \cite{fankhauser2018}.
Such methods run at 20-100Hz, and have been used for legged locomotion and footstep planning \cite{mastalli2017}.
One shortcoming of geometric mapping techniques is that they fail to account for terrain properties, which are an essential component in the stability of a legged robotic platform \cite{neunert2018}.
To compliment geometric mapping techniques, it is important to include information on terrain properties via semantic mapping.

\subsection{Semantic Mapping}

Semantic mapping is a broad field since high-level labels are task dependent.
Methods predicting terrain properties are of interest for robotic navigation, and we limit our focus within semantic mapping to such methods.
Semantic labels or attributes can be represented in abstract topological layers \cite{kuipers1991}, but it has become commonplace for semantic information to be estimated from images or point clouds using neural networks and then projected onto geometric representations \cite{gan2021}.
Visual information has been previously applied to predict the expected value for the coefficient of friction on roads \cite{wang2019} and other common terrain types \cite{brandao2016, noh2021}, as well as slip predictions for wheeled robots \cite{angelova2006}.
Current semantic mapping methods are non-recursive, meaning they cannot use priors to refine estimates, and as such, these methods are not robust while dealing with noisy sensor data.

One common semantic mapping paradigm is traversability estimation.
Traversability mapping has been used to bypass the need to estimate terrain properties by instead estimating which regions in the environment a robot can traverse \cite{gan2021, papadakis2013}.
Such mapping strategies often fail to consider that traversability is a function of the robot's internal state, such as the acceleration of the robot's center-of-mass or wheel velocity, and as such, traversability estimation methods often over- or under-approximate traversable regions \cite{kim2006}.
Recent work has shown how to incorporate traversibility classifications into a voxel occupancy grid map using a Bayesian inference framework to update traversibility estimates based on new observations \cite{gan2021}.
While this represents a large step towards a recursive framework for semantic mapping, it fails to address the short-comings associated with the robot-dependent nature of traversability.

\section{Preliminaries} \label{sec:prelim}
This section introduces notation, geometric concepts, coordinate frames, semantic segmentation, and probability theory used within this paper.
% \subsection{Notation} \label{subsec:notation}
Vectors, written as columns, are typeset in bold and lowercase, while sets and matrices are typeset in uppercase. 
The element $i$ of a vector $\boldsymbol{x}$ is denoted as $x_i$.
An n-dimensional open (resp. closed) interval is denoted by $(a, b)^n$ (resp. $[a, b]^n$).
The power set of a set $\mathcal{A}$ is denoted as $2^{\mathcal{A}}$.
Throughout the paper we let $f$ refer to a probability mass or density function.

\subsection{Geometry} \label{subsec:geometry}
\vspace{-1mm}
Barycentric coordinate systems specify the location of a point with respect to the vertices of a b-dimensional simplex.
Given a point  $\boldsymbol{p} \in \mathbb{R}^a$ in Euclidean space, we compute the corresponding b-dimensional Barycentric coordinate representation $\boldsymbol{\lambda} \in \mathbb{R}^b$ using a change of basis function $g:\mathbb{R}^a \to \mathbb{R}^b$ \cite{warren2007}, with $a \geq b$.
We test whether a point $\boldsymbol{p}$ lies within a b-dimensional simplex using the Barycentric coordinates by the following theorem:

\begin{thm}[{\cite[(2)]{zhang2014}}]\label{thm:intriangle}
Let $\boldsymbol{p} \in \mathbb{R}^a$ be a point in Euclidean space and let $\boldsymbol{\lambda} \in \mathbb{R}^b$ be its corresponding Barycentric coordinates.
Point $\boldsymbol{p}$ lies within the simplex if and only if  $\lambda_i \in (0, 1)$ for all $\lambda_i$ in $\boldsymbol{\lambda}$.
\end{thm}

\subsection{Sensors and Coordinate Frames} \label{subsec:coordframe}
\vspace{-1mm}
To simplify exposition, we assume the following:

\begin{assum}
    An RGB-D camera that is attached to a robot with known camera intrinsics is used to collect data.
\end{assum}
These RGB-D images, camera intrinsics, and an estimated camera pose are used to compute projected point clouds within the camera frame.
This process is discussed here as well as in Section \ref{subsec:semseg}.
This subsection introduces the notation used to describe coordinate frames and transformations between coordinate frames.
Note that one could extend the algorithms presented in this paper to multiple cameras or even LiDAR; we focus on a single RGB-D camera for simplicity.

We denote a vector to point $\boldsymbol{p}$ in coordinate frame $A$ as $\boldsymbol{p}_A$.
The rotation matrix from coordinate frame $A$ to $B$ parameterized by the rotation angles $\boldsymbol{q}$ between the two frames is denoted as $R_A^B(\boldsymbol{q})$.
Three coordinate frames are used within the paper: the world frame $W$, the sensor frame $S$, and the mapping frame $M$.
The inertial frame is space-fixed and the environment is assumed to be static with respect to this frame. 
The sensor frame $S$ is located at the center of the camera with the z-axis pointed out of the camera into the scene.
The transformation from the robot's center-of-mass to the sensor frame is static, and we assume that this transformation is known.
Lastly, the mapping frame $M$ is defined in relation to the location of the robot.
Its origin corresponds to the robot's center-of-mass projected onto the terrain, the x-axis (resp. y-axis) is oriented towards the front (resp. left-side) of the robot, and the z-axis is aligned with the z-axis of the inertial frame.
Given the vector to any range sensor measurement point in the sensor frame, $\boldsymbol{p}_S$, we transform it into the mapping frame via an affine transform:
\begin{equation} \label{eq:maptrans}
    \boldsymbol{p}_{M} = R_{S}^{M^\intercal} (\boldsymbol{q})\cdot \boldsymbol{p}_S - \boldsymbol{t}_S^M
\end{equation}
where $\boldsymbol{t}_S^M$ represents the translation from the sensor frame to the map frame.

\begin{figure*} [!t]
    \centering
    \includegraphics[width=\textwidth, height=5.5cm]{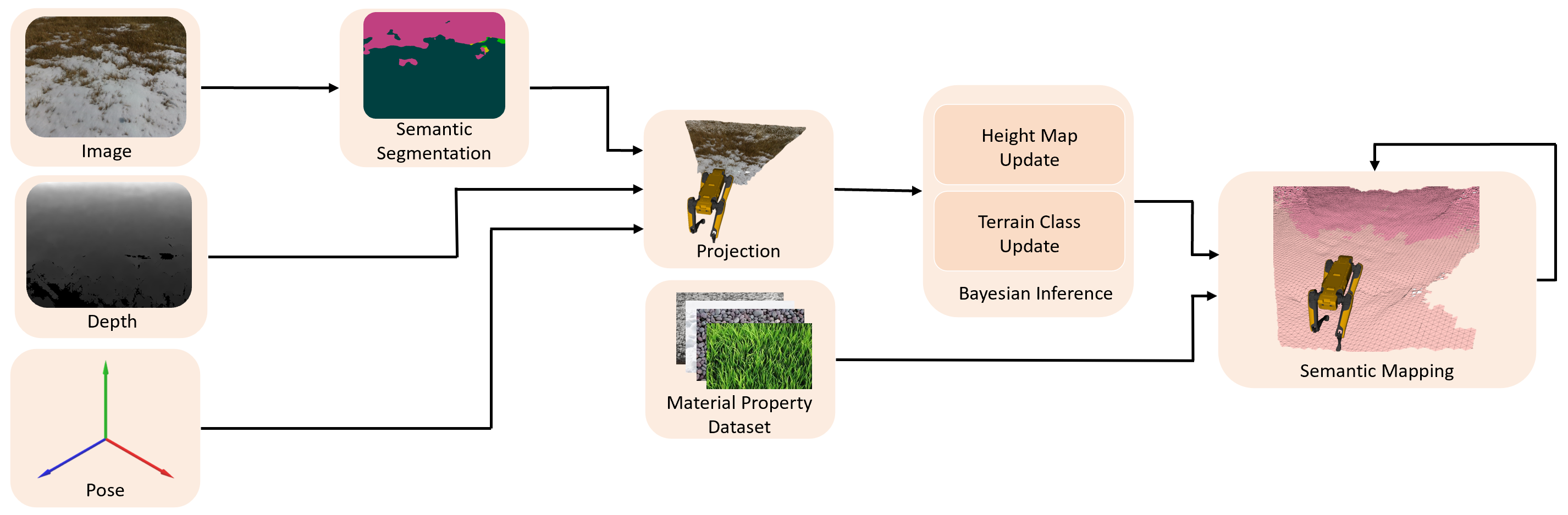}
    \caption{A flow diagram illustrating the behavior of Algorithm \ref{alg:mapping}. 
    RGB-D images are semantically segmented using an off-the-shelf semantic segmentation network.
    Using the camera's estimated pose and associated depth image, the pixel-wise probabilistic terrain class estimates are projected into the map. 
    The height map is updated using a 1D Kalman filter and the terrain class estimates, alongside our novel material property dataset, are used to recursively estimate terrain properties for each region of the map.}
    \label{fig:flow_diagram}
\end{figure*}

\subsection{Semantic Segmentation} \label{subsec:semseg}
\vspace{-1mm}
Semantic segmentation assigns class probability scores to each pixel in an image.
The classes in semantic segmentation are task-dependent.
This paper focuses on the list of terrain classes described in Table \ref{table:friction}.

It is common to use convolutional neural networks to estimate pixel-wise class probability scores  \cite{garcia2018}.
Let $I \in \mathbb{R}^{w \times h \times 3}$ denote an RGB image, where $h,w$ are the height and width of the image in pixels.
A trained semantic segmentation network takes an input image and outputs the pixel-wise terrain class probability scores $T \in \mathbb{R}^{w \times h \times k}$ for $k$ terrain classes in the form of a k-dimensional Categorical Distribution.
The accuracy of the semantic segmentation depends on the network used. 
Note, this is not the emphasis of this paper.
We use the aligned depth image $D \in \mathbb{R}^{w \times h}$ to project the pixel-wise terrain class probability scores $T$ into the sensor frame using the camera intrinsics and the camera projection equation \cite[(10.38)]{nixon2012} to obtain a point cloud representation of the semantically segmented image.

\subsection{Probability} \label{subsec:probability}
\vspace{-1mm}
This subsection reviews the Categorical Distribution, the Dirichlet Distribution, and their relation to one another.
We denote a random variable as $z$ or the vector of random variables as $\boldsymbol{z}$.
The Categorical Distribution is a discrete k-dimensional distribution parameterized by a vector $\boldsymbol{\theta} \in [0, 1]^k$.
The probability mass function of the Categorical Distribution represents the probability that sample $z$ belongs to class $i$, where $i \in \{1, 2, \dots, k\}$:
\begin{equation} \label{eq:categorical}
    f(z = i | \boldsymbol{\theta}) = \theta_i
\end{equation}

The Dirichlet Distribution is a continuous k-variate probability distribution which is parameterized by a vector $\boldsymbol{\alpha} \in \mathbb{R}^k_{\geq 0}$ of positive reals.
The probability density function of the Dirichlet Distribution is defined below:
\begin{equation} \label{eq:dirichlet}
    f(\boldsymbol{\theta} | \boldsymbol{\alpha}) = \frac{\Gamma(\sum_{j=1}^k \alpha_j)}{\sum_{j=1}^k \Gamma(\alpha_j)} \prod_{j=1}^k \theta_j^{\alpha_j-1}
\end{equation}
where 
\begin{equation}
    \Gamma(\alpha_j) = \int_0^\infty x^{\alpha_j-1} \exp(-x) dx.
\end{equation}

Suppose we obtain $n$ measurements $\mathcal{Z} = \{z_1, \dots, z_n\}$ of a given region, represented as random variables drawn from a Categorical Distribution.
Our goal is to apply Bayesian inference to predict the probability that a new measurement of the same region belongs to terrain class $i$ given prior measurements $\mathcal{Z}$.
That is we want to compute $f(z = i| \mathcal{Z}, \boldsymbol{\alpha})$.
Note, that we have assumed for full generality that $f(z| \mathcal{Z}, \boldsymbol{\alpha})$ is a function of some hyperparameters $\boldsymbol{\alpha}$.
To do this, one could compute $f(z| \mathcal{Z}, \boldsymbol{\alpha}) = \int_{\boldsymbol{\theta}} f(z|\boldsymbol{\theta}) f(\boldsymbol{\theta}|\mathcal{Z},\boldsymbol{\alpha}) \text{d} \theta$, but this would require constructing $f(\boldsymbol{\theta}| \mathcal{Z}, \boldsymbol{\alpha})$. 
By applying Bayes Theorem, one can write
\begin{align} \label{eq:bayes_integral}
    f(z| \mathcal{Z},\boldsymbol{\alpha}) &= \int_{\boldsymbol{\theta}} f(z|\boldsymbol{\theta}) \frac{f(\mathcal{Z}|\boldsymbol{\theta},\boldsymbol{\alpha}) f(\boldsymbol{\theta}|\boldsymbol{\alpha})}{f(\mathcal{Z}|\boldsymbol{\alpha})} \text{d}\theta.
\end{align}
Generally, this integral is hard to compute exactly. 

To compute a closed form expression for $f(z|\mathcal{Z},\boldsymbol{\alpha})$, we use the notion of conjugate prior \cite{tu2014}.
In particular, we choose to represent $f(\boldsymbol{\theta}|\boldsymbol{\alpha})$ as a Dirichlet Distribution, which is the conjugate prior to the Categorical Distribution $f(z|\boldsymbol{\theta})$.
With this choice, one can prove that $f(\boldsymbol{\theta}| \mathcal{Z}, \boldsymbol{\alpha})$ is also Dirichlet Distribution parameterized by a vector $\Tilde{\boldsymbol{\alpha}}$:
\begin{align} \label{eq:dirichlet_posterior}
    \tilde{\alpha}_j &= \alpha_j + \sum_{z_i \in \mathcal{Z}} 1\{z_i = j\}, 
\end{align}
where $1\{z_i = j\}$ is equal to $1$ when the expected terrain class of measurement $z_i$ is class $j$ and is zero otherwise \cite{tu2014}.
By using this property in \eqref{eq:bayes_integral}, one can prove \cite[(3)]{tu2014} the probability that a new measurement of the same region belongs to terrain class $i$ given prior measurements $\mathcal{Z}$ is:
\begin{align}
\label{eq:dir_to_cat}
    f(z = i| \mathcal{Z},\boldsymbol{\alpha}) = \frac{\tilde{\alpha}_i}{\sum_{j=1}^k \tilde{\alpha}_j}.
\end{align}

\section{Semantic Mapping and Bayesian Inference} \label{sec:method}
\begin{algorithm}[t]
\DontPrintSemicolon

\BlankLine
  
\Algorithm{}
$\mathcal{G} \leftarrow$ groundPlane() \tcp{Sec. \ref{subsec:mesh_initialization}}\label{alg:line:ground}
$\mathcal{V}$ collection of vertices
\label{alg:line:vertexinit} \tcp{Sec. \ref{subsec:mesh_initialization}}
$\Xi \leftarrow$ triangulation($\mathcal{V}$) \label{alg:line:faceinit} \tcp{Sec. \ref{subsec:mesh_initialization}}
 \While{robot is running}
 {
 $I, D, \boldsymbol{q} \leftarrow$ getImageAndSensorPose()  \label{alg:line:sensorinfo} \;
  $T \leftarrow$ semanticallySegmentImage($I$) \label{alg:line:semseg}\;
 $P_M \leftarrow$ projectImage($T, D, \boldsymbol{q}$) \label{alg:line:project} \tcp{Sec. \ref{subsec:semseg}}
assign points $\boldsymbol{p}_M \in P_M$ to $\xi \in \Xi$  \label{alg:line:assign} \tcp{Alg. 2}
$\mathcal{V} \leftarrow$ updateElevationMap($\mathcal{V}, \Xi$)  \label{alg:line:elmap} \tcp{Alg. 3}
$\Xi \leftarrow$ updateTerrainPrediction($P, \Xi$)  \label{alg:line:semmap} \tcp{Sec. \ref{sec:terrain_propety_est}}}
\caption{Recursive Semantic Mapping} \label{alg:mapping}
\end{algorithm}

As illustrated in Fig. \ref{fig:flow_diagram}, this section summarizes our robot-centric semantic mapping algorithm used to estimate the terrain surface profile and properties using a triangular mesh representation given an RGB-D camera with known pose (Algorithm \ref{alg:mapping}).
Subsequent sections describe each step of Algorithm \ref{alg:mapping} in detail.
The mesh is described using two collections.
The first is the collection $\mathcal{V} \subset (\mathbb{R}^4)^m$ of vertices $\boldsymbol{v} = [v_x, v_y, v_z, v_{\sigma^2}]$, where $m$ is the number of vertices within the mesh.
The first three components of a vertex, $v_x$, $v_y$, and $v_z$, correspond to the Euclidean position of the vertex with respect to the mapping frame $M$, and the last component $v_{\sigma^2}$ corresponds to the variance of $v_z$.
The second is the collection $\Xi \subset \mathcal{V}^3 \times (\mathbb{R}^{3+k})^l \times \mathbb{R}^k_{\geq0}$ of mesh elements, or faces, $\xi$.
An element $\xi$ is a collection of three components: the three vertices whose connecting line segments define the perimeter of the face, interior points, and a vector of Dirichlet parameters.
The interior points are discussed in Section \ref{subsec:point_assignment}.

We start by defining a flat ground plane $\mathcal{G}$ with zero height (Line \ref{alg:line:ground}).
Next, vertices $\boldsymbol{v} \in \mathcal{V}$ and mesh elements $\xi \in \Xi$ are initialized (Lines \ref{alg:line:vertexinit}-\ref{alg:line:faceinit}, Section \ref{subsec:mesh_initialization}).
% We retrieve the RGB-D image and camera pose estimate from the robot, where $I$ denotes the RGB image, $D$ denotes the depth image, and $\boldsymbol{q}$ denotes the sensor pose in the world frame (Line \ref{alg:line:sensorinfo}).
We retrieve the RGB-D image, $I$ and $D$, and camera pose estimate in the world frame, $\boldsymbol{q}$, from the robot (Line \ref{alg:line:sensorinfo}).
A semantic segmentation network takes the RGB image $I$ and outputs pixel-wise terrain class probability scores, $T$ (Line \ref{alg:line:semseg}).
These pixel-wise scores are projected into the mapping frame (Line \ref{alg:line:project}, Section \ref{subsec:coordframe}) and assigned as interior points to a mesh element $\xi$ (Line \ref{alg:line:assign}, Section \ref{subsec:point_assignment}).
Interior points are used to compute vertex heights, vertex height covariance, and terrain labels of the corresponding mesh element $\xi$. 
The height map is updated using the projected points (Line \ref{alg:line:elmap}, Section \ref{subsec:elevation_map}). and terrain properties are recursively updated via the Dirichlet-Categorical conjugacy relationship (Line \ref{alg:line:semmap}, Section \ref{sec:terrain_propety_est}).

\section{Recursive Elevation Mapping} \label{sec:recursive_elevation_mapping}
This section describes how our algorithm recursively estimates the elevation map given range sensor measurements.
We begin with the initialization of a piece-wise planar triangular mesh that represents the contact surface.
Next we construct a technique to assign range sensor measurements as interior points to their corresponding triangular mesh element $\xi$.
Finally, we describe how to update the elevation map.

\subsection{Mesh Initialization} \label{subsec:mesh_initialization}
\vspace{-1mm}
At startup, we define a flat ground plane $\mathcal{G}$ with zero height (Line \ref{alg:line:ground}) and initialize a grid pattern of evenly-spaced vertices $\boldsymbol{v} \in \mathcal{V}$ with zero height and zero variance (Line \ref{alg:line:vertexinit}).
The set of faces $\xi \in \Xi$ are initialized (Line \ref{alg:line:faceinit}) by triangulating these vertices into a set of equal-sized, isosceles, right-angled triangles.
The set of interior points of each face is initialized as an empty set and the Dirichlet parameters are initialized as a vector of zeros.

\subsection{Point Assignment} \label{subsec:point_assignment}
\vspace{-1mm}
Interior points represent the set of points $\boldsymbol{p}_M$ whose projection lie within the 2-dimensional simplex defined by the perimeter of $\xi$.
The process by which points are projected and assigned as interior points is described in Alg. \ref{alg:assigntoelems}.
The camera on the robot collects RGB-D images that are semantically segmented using a neural network (Section \ref{subsec:semseg}) before being projected into the mapping frame as a point cloud $P_M \subset (\mathbb{R}^{3 + k})^n$ (Lines \ref{alg:line:sensorinfo}-\ref{alg:line:project}, Alg. \ref{alg:mapping}) made up of $n$ points (Section \ref{subsec:coordframe}).
The first three components of a point $\boldsymbol{p}_M \in P_M$ correspond to the Euclidean coordinates of the point in the mapping frame, while the last $k$ components correspond to the terrain class probability score output from the semantic segmentation network.
Next, we project points $\boldsymbol{p}_M \in P_M$ and the mesh vertex coordinates $\boldsymbol{v} \in \mathcal{V}$ onto the ground plane $\mathcal{G}$ by projecting $\boldsymbol{p}_M$ and $\boldsymbol{v}$ onto their first two coordinates.
We obtain the projected point $\Tilde{\boldsymbol{p}}_M=[p_x, p_y]$ as well as the three projected vertices $\Tilde{\boldsymbol{v}}_i = [v_{xi}, v_{yi}]$ of a mesh element $\xi$ for each $i \in  \{1, 2, 3\}$.
Given $\Tilde{\boldsymbol{p}}$ and $\Tilde{\boldsymbol{v}}$, we compute the Barycentric coordinates $\boldsymbol{\lambda} = [\lambda_1, \lambda_2, \lambda_3]$ using the following linear transform (Line \ref{alg:line:bary_trans}, Alg. \ref{alg:assigntoelems}):
\begin{equation}
    \begin{bmatrix}
        \lambda_1 \\
        \lambda_2 \\
        \lambda_3
    \end{bmatrix} =
    \begin{bmatrix}
        1 & 1 & 1\\
        v_{x1} & v_{x2} & v_{x3} \\
         v_{y1} & v_{y2} & v_{y3}
    \end{bmatrix}^{-1} 
    \begin{bmatrix}
        1 \\
        p_x \\
        p_y
    \end{bmatrix}.
\end{equation}
We apply Theorem \ref{thm:intriangle} to determine whether to assign a point as an interior point to mesh element $\xi$ (Line \ref{alg:line:interior_point}, Alg. \ref{alg:assigntoelems}).

\begin{algorithm}[t]
\DontPrintSemicolon
\Requires{$P_M$, $\mathcal{V}$, $\Xi$}

\For{$\boldsymbol{p}_M \in P_M$}{
    \For{$\xi \in \Xi$}{
        $\Tilde{\boldsymbol{p}}_M, \Tilde{\boldsymbol{v}} \leftarrow$ groundPlaneProjection($\xi, \boldsymbol{p}_M)$ \;
        $\boldsymbol{\lambda} \leftarrow$ computeBarycentricCoords($\Tilde{\boldsymbol{p}}_M, \Tilde{\boldsymbol{v}}$) \label{alg:line:bary_trans} \;
        \If{ for all $\lambda_i \in \boldsymbol{\lambda}, \lambda_i \in [0,1]$}{
            $\xi \leftarrow$ add interior point $\boldsymbol{p}_M$  \label{alg:line:interior_point}
        }
    }
}

\caption{Assign points $\boldsymbol{p}_M \in P_M$ to $\xi \in \Xi$} \label{alg:assigntoelems}
\end{algorithm}

\subsection{Elevation Map Computation} \label{subsec:elevation_map}
\vspace{-1mm}
Next we describe how Algorithm \ref{alg:elevation} recursively estimates the elevation map given the range sensor measurements (Line \ref{alg:line:elmap}, Alg. \ref{alg:mapping}).
These interior points from the preceding section are now used to update the elevation map.

For a vertex $\boldsymbol{v} \in \mathcal{V}$, we take the interior points from the surrounding mesh elements (Line \ref{alg:line:surroundinginterior}, Alg. \ref{alg:elevation}) and apply a 1-dimensional Kalman filter update to estimate the mean height $v_z$ and variance $v_{\sigma_z}$ of the vertex.
Given the depth image used to compute the point cloud $P_M$ has sensor noise, there is variance in the Euclidean coordinates of $\boldsymbol{p}_M \in P_M$.
Once assigned to a mesh element, the elevation map depends only on the height of the points $\boldsymbol{p}_M \in P_M$, so we only consider the variance of the third Euclidean coordinate.

Recall that the third component of $\boldsymbol{p}_M$, which we denote $p_{M,3}$ describes its height. 
By the error propagation law \cite{soler2002}, the variance of $p_{M,3}$ is computed (Line \ref{alg:line:heightandvar}, Alg. \ref{alg:elevation}):
\begin{equation} \label{eq:variance}
    \sigma^2 = J_s \Sigma_s J_s^\intercal + J_p\Sigma_p J_p^\intercal,
\end{equation}
where $\Sigma_s$ and $\Sigma_p$ are the range sensor measurement noise and the sensor pose covariance matrix, respectively, and $J_s$ and $J_p$, are constructed by taking the following partial derivatives:
\begin{equation}
    J_s := \frac{\partial p_{M,3} }{\partial \boldsymbol{p}_{\mathcal{S}}} = (R_{\mathcal{S}}^{\mathcal{M}\intercal}(\boldsymbol{q}))_3
\end{equation}
\begin{equation}
    J_p := \frac{\partial p_{M,3}}{\partial R_\mathcal{S}^\mathcal{M}(\boldsymbol{q})} = (R_\mathcal{S}^{\mathcal{M} \intercal}(\boldsymbol{q}))_3 \times \boldsymbol{p}_{\mathcal{S}},
\end{equation}
where $(R_\mathcal{S}^{\mathcal{M} \intercal}(\boldsymbol{q}))_3$ denotes the third row of $R_\mathcal{S}^{\mathcal{M} \intercal}(\boldsymbol{q})$ and $\times$ denotes the cross product. 
The mean and variance of the vertex height, $v_z$ and $v_{\sigma^2}$, are updated using a 1-dimensional Kalman filter (Line \ref{alg:line:1DKalmanUpdate}, Alg. \ref{alg:elevation}) for all the interior points from the surrounding mesh elements:
\begin{equation}
    v_z \leftarrow \frac{v_z \cdot \sigma^2 + z \cdot v_{\sigma^2}}{\sigma^2 + v_{\sigma^2}}
\end{equation}
\begin{equation}
    v_{\sigma^2} \leftarrow \frac{v_{\sigma^2} \cdot \sigma^2}{v_{\sigma^2} + \sigma^2}.
\end{equation}

% \vspace{-3.5mm}

\begin{algorithm}[t]
\DontPrintSemicolon
% \LinesNotNumbered
\Requires{$\mathcal{V}$, $\Xi$}

$\boldsymbol{q} \leftarrow$ cameraPose() \; 

$\Sigma_s \leftarrow$ sensorNoiseModel() \;

$\Sigma_p \leftarrow$ robotPoseCovariance() \;

\For{$\boldsymbol{v} \in V$}{

        $\Bar{P}_M \leftarrow$ getSurroundingInteriorPoints($\boldsymbol{v}, \Xi$) \label{alg:line:surroundinginterior} \;
        
        \For{$\Bar{\boldsymbol{p}}_M \in \Bar{P}_M$}{
        
            $\bar{p}_{M,3}, \sigma^2 \leftarrow$ heightVariance($\Bar{\boldsymbol{p}}_M, \boldsymbol{q}, \Sigma_s, \Sigma_p$) \label{alg:line:heightandvar}\;
        
            $v_z, v_{\sigma^2} \leftarrow$ 1DKalmanFilter($\bar{p}_{M,3}, \sigma^2$) \label{alg:line:1DKalmanUpdate}
        
        }
}

\Return{$\mathcal{V}, \Xi$}

\caption{Update Elevation Map} \label{alg:elevation}
\end{algorithm}

\section{Recursive Terrain Property Estimation} \label{sec:terrain_propety_est}
% \vspace{-2mm}
The objective of our semantic mapping algorithm is to estimate the distribution of terrain properties of the environment around the robot.
Motivated by prior work \cite{nguyen2021, angelova2006}, we use data to construct a conditional probability distribution, $ f(\boldsymbol{\psi} \mid z = i)$, of terrain property, $\boldsymbol{\psi}$, conditioned on a terrain class estimate for a region $z = i$. 
Using this model, given $\mathcal{Z}$ measurements of a region that is interior to  $\xi$, we then apply the Law of Total Probability to compute this region's predicted terrain property:
\begin{equation} \label{eq:contitionalindep_infer}
        f(\boldsymbol{\psi} \mid \mathcal{Z}, \boldsymbol{\alpha}) = \sum_{i = 1}^k f(\boldsymbol{\psi} \mid z = i) f(z = i \mid \mathcal{Z}, \boldsymbol{\alpha})
\end{equation}

\noindent Note, this paper is interested in estimating the friction coefficient; however, the presented theory can be extended to other terrain properties of interest.
We next discuss the components of \eqref{eq:contitionalindep_infer} before presenting a closed-form solution for recursively estimating the coefficient of friction within our semantic map (Line \ref{alg:line:semmap}, Alg. \ref{alg:mapping}).

% \subsection{Recursive Terrain Class Inference} \label{subsec:terrain_class_inference}
% \vspace{-1mm}
Following Section \ref{subsec:point_assignment}, semantically segmented pixels $\mathcal{Z}$ are projected into the mapping frame and assigned as interior points to mesh elements $\xi$.
Recall from Sections \ref{subsec:semseg}, \ref{subsec:probability} and \eqref{eq:categorical}, the pixel-wise terrain class probability generated from a semantic segmentation network represent parameters $\boldsymbol{\theta}$, which are used to update $\boldsymbol{\alpha}$ via \eqref{eq:dirichlet_posterior}.
For each mesh element, we compute $f(z = i \mid \mathcal{Z}, \boldsymbol{\alpha})$ using \eqref{eq:dir_to_cat}.

% \subsection{Terrain Properties from Terrain Classes} \label{subsec:prop_from_class}
% \vspace{-1mm}
Terrain properties are not constant across a terrain class and thus should not be estimated by a single value.
Rather, these properties should be modelled using a conditional probability distribution $f(\boldsymbol{\psi} \mid z = i)$.
This model is fit using data collected from each class.
As we show in Section \ref{sec:implementation}, we create a well-fit model by selecting an appropriate mean $\mu_i$ and variance $\sigma_i^2$ for a unimodal Gaussian distribution $f(\boldsymbol{\psi}\mid z=i)=\mathcal{N}(\mu_i, \sigma_i^2)$.

% \subsection{Closed-Form Terrain Property Inference} \label{subsec:terrain_inf}
Substituting \eqref{eq:dir_to_cat} and the formula for the unimodal Gaussian into \eqref{eq:contitionalindep_infer} gives a closed-form estimate for the terrain properties within a mesh element:
\begin{equation} \label{eq:property_comp}
    f(\boldsymbol{\psi} \mid \mathcal{Z}, \boldsymbol{\alpha}) = \sum_{i=1}^k \frac{\alpha_i}{\sum_{j=1}^k \alpha_j} \mathcal{N}(\mu_i, \sigma_i^2).
\end{equation}
This is a multimodal Gaussian distribution where each mode is weighted relative to the recursively updated terrain class likelihood.
Note that \eqref{eq:property_comp} can be extended to use terrain property models other than the unimodal Gaussian distribution.

\section{Implementation} \label{sec:implementation}
This section describes the implementation of our algorithm.
Algorithm \ref{alg:mapping} is implemented in C++ and includes a Robot Operating System (ROS) interface\footnote[1]{\href{https://github.com/roahmlab/sel_map}{https://github.com/roahmlab/sel\_map}}.
Our implementation features noise models for the Realsense RGB-D camera and an interface to include additional sensor noise models.
We evaluated our method on a desktop with a 3.1GHz Ryzen 3600 processor, 32GB of RAM and an Nvidia RTX 2080 Ti GPU.

We use \eqref{eq:contitionalindep_infer} to estimate terrain properties from semantically segmented RGB-D images.
This requires a model relating terrain class to terrain properties.
The dataset published in \cite{panahandeh2017} is insufficient to compute a probabilistic model as it only contains approximately three friction measurements per terrain class, and neither the Gaussian friction model proposed in \cite{noh2021} nor their friction data is currently publicly available.
To compute a probabilistic model we introduce a novel dataset of friction measurements across ten terrain classes and make this data publicly accessible.\footnote[2]{\href{https://github.com/roahmlab/terrain_friction_dataset}{https://github.com/roahmlab/terrain\_friction\_dataset}}
We discuss the steps for data collection and subsequent model fitting in the following paragraphs.

Our primary focus in this paper is on estimating the coefficient of friction.
We built a device to measure the coefficient of friction using the pulling force measured using a load cell, the known weight of the device, and $g = 9.81 \frac{m}{s^2}$:
\begin{equation}
    \mu = \frac{F_{pull}}{mg}.
\end{equation}
\noindent Approximately ten thousand data samples were collected and the data was post-processed using a low-pass filter to remove measurement noise from the load cell.
To model $ f(\boldsymbol{\psi}\mid z=i)$, we fit the unimodal Gaussian, Weibull, and log-normal distributions, and we assessed the goodness-of-fit for each distribution using the Kolmogorov-Smirnov test \cite{massey1951}.
The unimodal Gaussian distribution had the highest average Kolmogorov-Smirnov score across all terrain classes demonstrating that the Gaussian model generalized the best over the entire dataset.
We therefore use the unimodal Gaussian model to model $f(\boldsymbol{\psi} \mid z = i)$.
Table \ref{table:friction} contains the mean and variance parameters of each unimodal Gaussian distribution for each terrain class of interest.

\section{Results} \label{sec:results}
We evaluate the performance of our mapping framework in the Carla simulation environment \cite{dosovitskiy2017} and on a physical legged robot.
In simulation, we compare our method against two baselines representing state-of-the-art terrain property estimation methods and illustrate that our method outperforms both baselines.
We also demonstrate our method in real-world indoor and outdoor environments on a quadruped robot and compare it to a state-of-the-art traversability estimation method.
A supplementary video demonstrates the proposed mapping framework on the Spot quadruped.

\subsection{Computational Performance Evaluation}
\vspace{-1mm}
We ran Alg. \ref{alg:mapping} using a $1$m$\times 1$m mesh and varied mesh element lengths with random input images and associated ground-truth semantic segmentations to evaluate the computational speed and memory requirements.
Approximately 45-55MB of memory is required to store the mesh.
With $1$cm mesh element lengths, Algorithm \ref{alg:mapping} takes 527ms to run, of which the semantic segmentation network from \cite{zhou2018} takes 477ms (Alg. \ref{alg:mapping} Lines \ref{alg:line:semseg}-\ref{alg:line:project}), and the elevation map and terrain property update takes 50ms (Alg. \ref{alg:mapping} Lines \ref{alg:line:assign}-\ref{alg:line:semmap}).
These computation times were computed by averaging across $300$ trials.
The computation time for semantic segmentation is network dependent, and using Fast-SCNN \cite{poudel2019} the total computation time is reduced to approximately $200$ms.
A thorough evaluation of the computational times for Algorithm \ref{alg:mapping} with two semantic segmentation networks, \cite{zhou2018} and \cite{poudel2019}, and varying mesh element lengths is given in Figure \ref{fig:comp_times}.

\begin{table}[!t]
\centering
\begin{tabular}{|c|cc|}
\hline
\multirow{2}{*}{Terrain Class} & \multicolumn{2}{c|}{Coefficient of Friction Gaussian Parameters} \\ \cline{2-3} & \multicolumn{1}{c|}{\; \; \; \; \; $\mu$ \; \; \; \; \;} & $\sigma$ \\ \hline
Concrete & \multicolumn{1}{c|}{0.543} & 0.065 \\ \hline
Grass & \multicolumn{1}{c|}{0.577} & 0.077 \\ \hline
Pebbles & \multicolumn{1}{c|}{0.428} & 0.059 \\ \hline
Rocks & \multicolumn{1}{c|}{0.478} & 0.113 \\ \hline
Wood & \multicolumn{1}{c|}{0.372} & 0.055 \\ \hline
Rubber & \multicolumn{1}{c|}{0.616} & 0.048 \\ \hline
Rug & \multicolumn{1}{c|}{0.583} & 0.068 \\ \hline
Snow & \multicolumn{1}{c|}{0.390} & 0.071 \\ \hline
Ice & \multicolumn{1}{c|}{0.192} & 0.046 \\ \hline
Laminated Flooring & \multicolumn{1}{c|}{0.311} & 0.045 \\ \hline
\end{tabular}
\caption{Unimodal Gaussian parameters computed from coefficient of friction data collected across multiple terrain classes.}
\label{table:friction}
\end{table} \label{subsec:performance}

\subsection{Simulation} \label{subsec:simresults}
\vspace{-1mm}
We evaluate our terrain property estimation method in the Carla simulation environment \cite{dosovitskiy2017} where ground truth terrain property information is provided on a per-class basis.
The ground truth distribution for the coefficient of friction for each class is a unimodal Gaussian using the coefficient of friction models computed in Section \ref{sec:implementation}.
Within Carla, we collect RGB-D information from a camera mounted on the front of a car.
To estimate terrain class, we use the pre-trained semantic segmentation network presented in \cite{zhou2018} and trained on the ADE20K dataset.

\begin{figure}[t]
    \centering
    \includegraphics[width=\columnwidth]{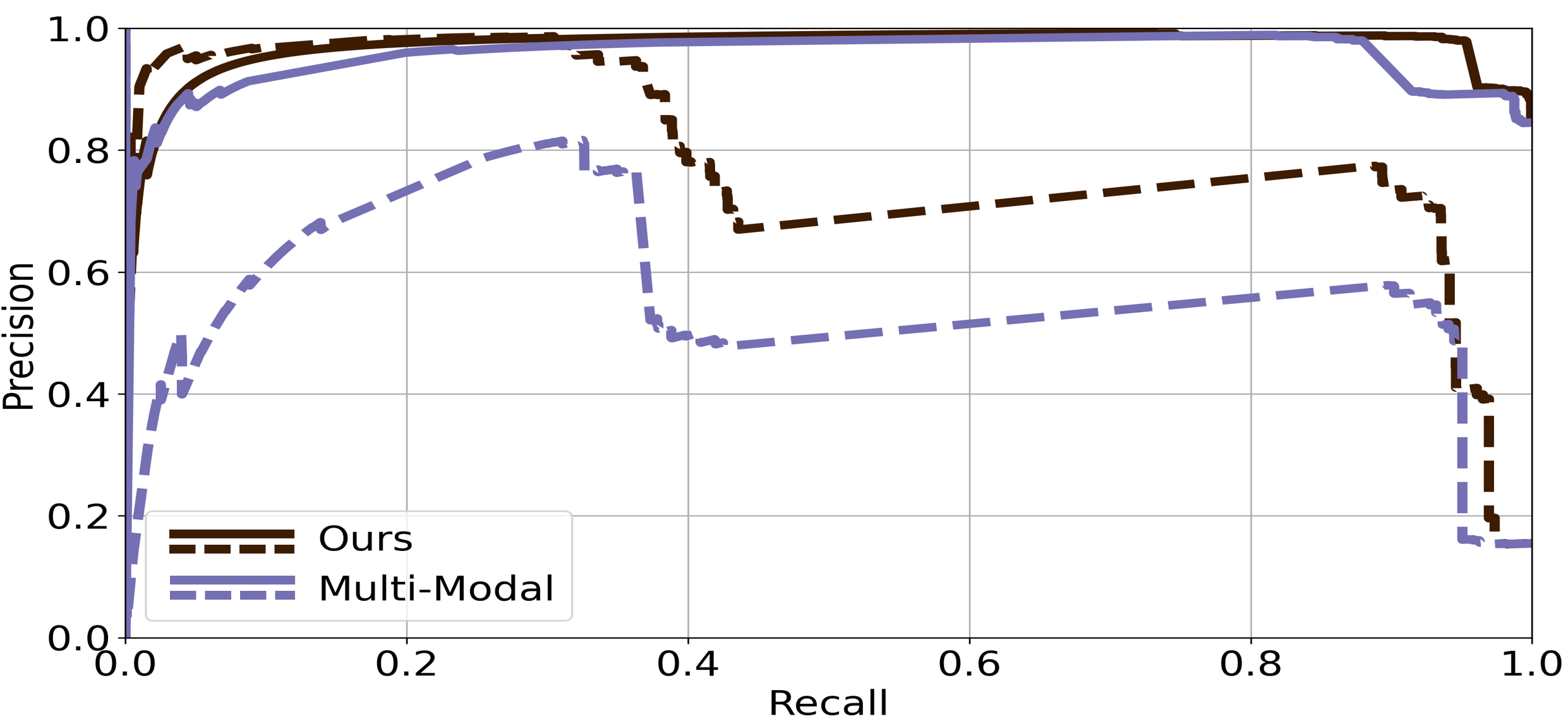}
    \caption{The Precision-Recall curve for terrain property estimation within the Carla simulator. We compare our method against the \textit{Multimodal Non-Recursive} baseline for regions of high friction coefficients ($\mu > 0.5$), plotted using a solid line, and low friction coefficients ($\mu \leq 0.5$), plotted using a dashed line. Our method’s performance is comparable to the baseline for regions of high friction, however, for regions of low friction our method significantly outperforms the baseline.
    }
    \label{fig:prcurve}
\end{figure}

\begin{figure}[!t]
    \centering
    \includegraphics[width=\columnwidth]{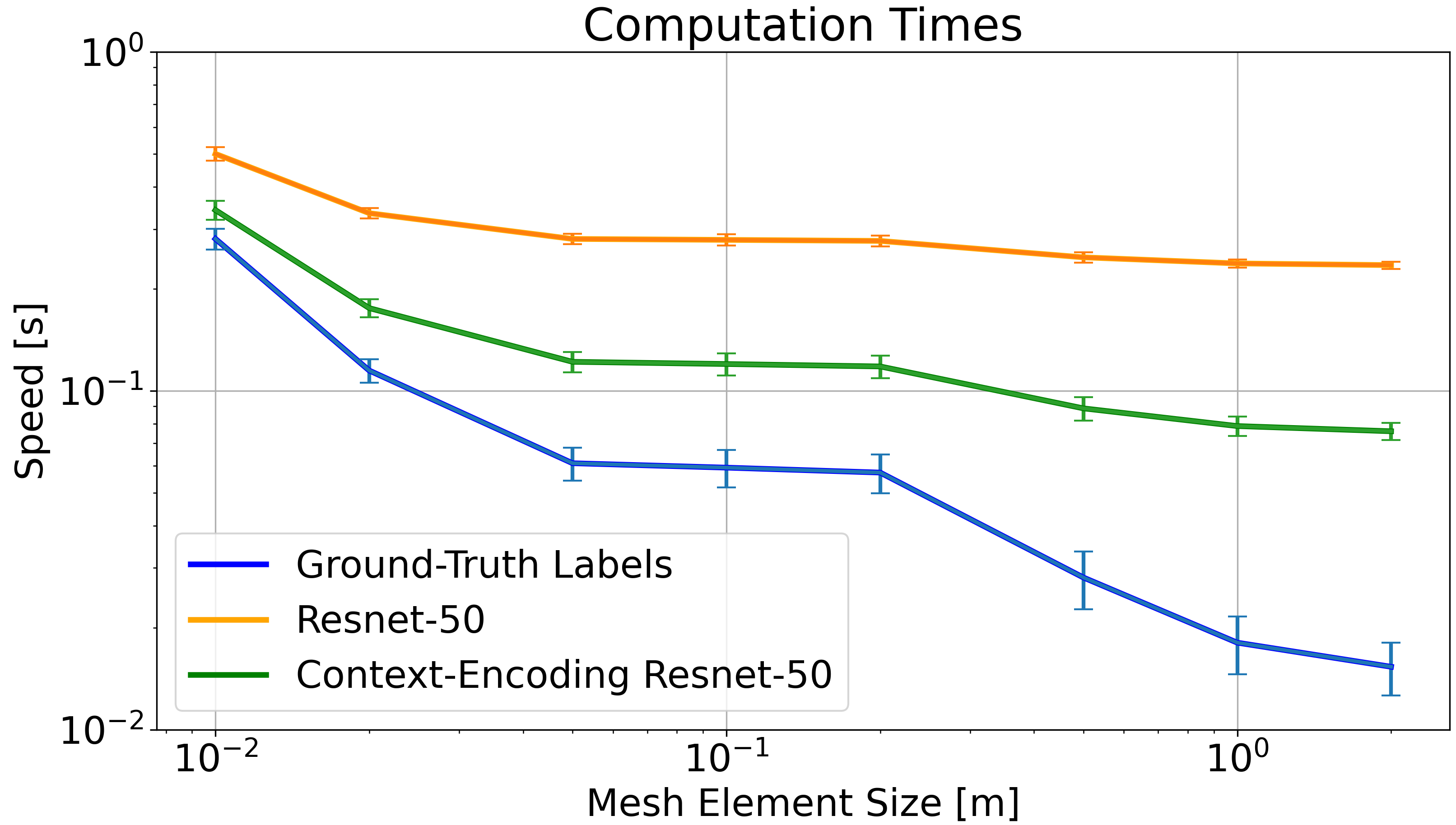}
    \caption{Computation times of Algorithm \ref{alg:mapping} for a $10$m $\times 10$m mesh with varying mesh element lengths using two off-the-shelf semantic segmentation networks, Resnet-50 \cite{zhou2018} and Context-Encoding Resnet-50 \cite{zhang2018}, as well as the baseline algorithm assuming ground-truth semantically segmented images. The ground-truth label experiments use a pre-generated semantic segmentation image, representing the speed of Algorithm \ref{alg:mapping} without considering the time required for semantic segmentation (Line \ref{alg:line:semseg}) and represents an lower bound on the speed of our algorithm. Error bars represent one standard deviation.}
    \label{fig:comp_times}
\end{figure}

We use coefficient of friction estimates to compare our method against two baselines representing the state of the art in the terrain property estimation literature. 
The first baseline, denoted as the \textit{Unimodal Non-Recursive} method, estimates the coefficient of friction by taking the most likely terrain class for a given mesh element at each instance in time and uses the unimodal Gaussian model with parameters from Table \ref{table:friction}.
This baseline is representative of methods such as those presented in \cite{wang2019, brandao2016, procopio2009}, which estimate the expected value of the coefficient of friction using convolutional neural networks. 
The second baseline, denoted as the \textit{Multimodal Non-Recursive} method, uses the full categorical distribution of a given semantically segmented mesh element to estimate the coefficient of friction as a multi-modal Gaussian distribution.
This equates to using \eqref{eq:property_comp} to compute the coefficient of friction directly from the pixel-wise categorical scores outputted from the semantic segmentation network.
This baseline is representative of the state of the art methods~\cite{angelova2006, noh2021} that use the terrain class to estimate terrain properties from RGB-D images.
These methods do not employ a recursive framework to update belief in terrain classifications.
We ran our algorithm and the baselines offline using the data collected within Carla and compared these estimates with the ground-truth distributions using a precision-recall curve (Fig. \ref{fig:prcurve}) and their Kullback-Leibler divergence scores (Table \ref{table:kld}).

Note, the lower the Kullback-Leibler divergence score, the more similar two distributions are.
From Table \ref{table:kld}, one can see that the first baseline performed poorly for the Carla dataset.
The score for the second baseline is lower than the first's, indicating the coefficient of friction distribution estimate of the second baseline is more similar to the ground-truth distribution.
Lastly, our proposed method performed best on the Kullback-Leibler divergence test and demonstrates that the coefficient of friction distribution estimated using our proposed method is the most similar to the ground-truth distribution.

\begin{table}[!t]
\centering
\begin{tabular}{|c|c|c|c|}
\hline
Method & KL Score ($\downarrow$) & \begin{tabular}[c]{@{}c@{}} Average\\ Precision\end{tabular} ($\uparrow$) & \begin{tabular}[c]{@{}c@{}} Average\\ Accuracy\end{tabular} ($\uparrow$)\\ \hline
\begin{tabular}[c]{@{}c@{}}Uni-Modal\\ Non-Recursive\end{tabular}   & 42.3 & 0.59 & 0.58 \\ \hline
\begin{tabular}[c]{@{}c@{}}Multi-Modal\\ Non-Recursive\end{tabular} & 3.7 & \textbf{0.99} & 0.93 \\ \hline
\textbf{Ours} & \textbf{2.4} & \textbf{0.99} & \textbf{0.95} \\ \hline
\end{tabular}
\caption{The Kullback-Leibler divergence scores, average precision, and average accuracy of the two baselines and our method when applied to the Carla simulation environment. Arrows depict whether a high ($\uparrow$) or low ($\downarrow$) score is desired. A bolded score indicates the best performing method in each criteria.}
\label{table:kld}
\end{table}

The precision-recall curve summarizes the trade-off between the true positive rate and the positive predicted value and is used to evaluate the performance of a multi-class classifier.
We use the average precision to evaluate the performance on the precision-recall curve as seen in Figure \ref{fig:prcurve}.
A higher average precision indicates a more accurate classifier.
For this evaluation, we divide the range of coefficient of friction values into low friction ($\mu \leq 0.5$) and high friction ($\mu > 0.5$) categories and compare the ability of our method and and the \textit{Multimodal Non-Recursive} method to correctly predict whether a given mesh element falls within the low or high friction category.
The results for \textit{Unimodal Non-Recursive} method is omitted due to poor performance.

Table \ref{table:kld} includes the performance of all methods using the three quantitative metrics.
Due to class imbalance within the simulation environment, more high-friction terrain classes are present in the data.
The baselines perform better for high-friction classes, but even with this class imbalance our method matches or outperforms both baselines across all evaluation criteria.
This shows our method is able to better predict the terrain friction properties than previous terrain estimation methods from the literature.

\subsection{Real-World} \label{subsec:realresults}
\vspace{-1mm}

\begin{figure}[!th]
  \centering
  \includegraphics[width=\linewidth]{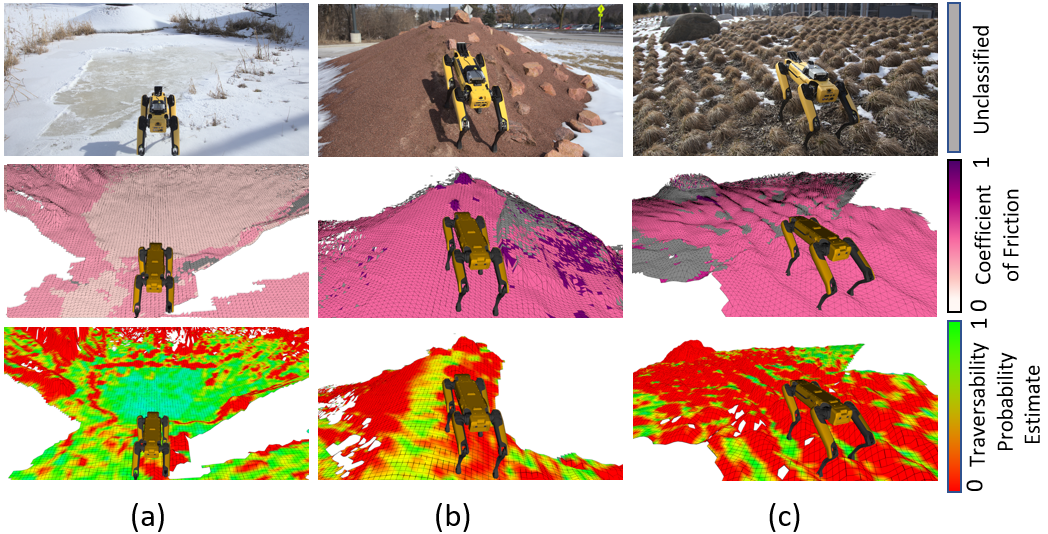}

  \caption{Each column depicts the performance of our proposed mapping algorithm (second row, the first column uses the Context-Encoding ResNet-50 trained on the Pascal dataset while the remaining columns use the RenNet-50 trained on ADE20K for terrain classification) when compared to a traversability estimation algorithm \cite{gan2021} (third row) applied on the scenes depicted in the top row. 
  Traversability estimation sometimes believes that a region is traversable when it is not, such as an icy surface (Fig.  5a). 
  In other scenarios, it believes that an area is intraversable when it is traversable such as on hills (Figs. 5b) and near low vegetation (Fig. 5c). 
  Our method makes no claims about traversability, instead it estimates the probability distribution of terrain properties for each mesh element along with the terrain geometry.}
  \label{fig:comparison}
\end{figure}
 
We ran our method on the Spot quadruped using an on-board Realsense D435 RGB-D camera.
Experiments were conducted both indoors and outdoors with a variety of terrain classes.
We compared our method to a state-of-the-art traversability mapping framework \cite{gan2021} to demonstrate the utility of our semantic mapping algorithm when compared to a traversability estimation algorithm.
Figure \ref{fig:comparison} illustrates the performance of both algorithms on a variety of examples across different terrains.
On an icy surface (Fig. 5a), for instance, our method is able to predict the low friction of the surface, while the traversability estimate assumes the surface is safe to walk on and provides no additional information regarding the surface.
Similarly, Figs. 5b and 5c illustrate that the traversability estimation incorrectly classifies regions which are traversable as intraversable while our method is able to predict the terrain geometry and properties.
When no terrain class from Table \ref{table:friction} is estimated within a mesh element, we make no friction estimate and color the mesh element grey (second row,  Fig. \ref{fig:comparison}).
Traversability depends on the means of robot locomotion and other robot-dependent factors.
In an effort to generalize, traversability estimation methods often over- or under-approximate traversable regions, supporting the conclusions reached by \cite{kim2006}.
In contrast, our method provides more information than binary traversability labels which better informs robots about their environment.

\section{Conclusions} \label{sec:conclusion}
We propose a Bayesian inference framework for real-time elevation mapping and terrain property estimation from RGB-D images.
Our method outperforms other algorithms both in simulation and the real-world.
Unlike traversability methods, our algorithm provides terrain property information that can enable robots to adjust their locomotion to traverse regions of low friction rather than just avoid them.

\addtolength{\textheight}{-1cm}   % This command serves to balance the column lengths

%%%%%%%%%%%%%%%%%%%%%%%%%%%%%%%%%%%%%%%%%%%%%%%%%%%%%%%%%%%%%%%%%%%%%%%%%%%%%%%%
%%%%%%%%%%%%%%%%%%%%%%%%%%%%%%%%%%%%%%%%%%%%%%%%%%%%%%%%%%%%%%%%%%%%%%%%%%%%%%%%

\bibliographystyle{IEEEtran}
\bibliography{references}

\end{document}